\documentclass[bimj,fleqn]{w-art}
\usepackage{times}
\usepackage{w-thm}
\usepackage[authoryear]{natbib}
\usepackage{mathrsfs}
\usepackage{graphicx}
\usepackage{algpseudocode}
\usepackage{amsfonts}
\usepackage{algorithm}
\usepackage{url}
\usepackage{xcolor}
\usepackage{afterpage}

\theoremstyle{plain}

\theoremstyle{definition}

\usepackage[]{graphicx}
\chardef\bslash=`\\ % p. 424, TeXbook

\begin{document}

	\Volume{}
	\Issue{}
	\Year{2024}
	\pagespan{1}{}
	  %%% semicolon and fullpoint added here for keyword style
	
	\title[Neural interval-censored survival regression with feature selection]{Neural interval-censored survival regression with feature selection}
	%% Information for the first author.
	\author[Meixide {\it{et al.}}]{Carlos García Meixide\footnote{Corresponding author: {\sf{e-mail: {carlos.garcia@icmat.es}}}}{\inst{1,2}}}
\address[\inst{1}]{{Instituto de Ciencias Matemáticas, CSIC}}
\address[\inst{2}]{{Departamento de Matemáticas, Universidad Autónoma de Madrid}}
	%%%%    Information for the second author
	\author[]{Marcos Matabuena\footnote{Corresponding author: {\sf{e-mail: mmatabuena@hsph.harvard.edu}}}\inst{3}}
	\address[\inst{3}]{Harvard University}
	%%%%    Information for the third author
	\author[]{Louis Abraham\inst{4}}
	\address[\inst{4}]{Université Paris 1 Panthéon-Sorbonne}

	\author[]{\\Michael R. Kosorok\inst{5}}
	%%%%    \dedicatory{This is a dedicatory.}
	\address[\inst{5}]{University of North Carolina, Chapel Hill}
	
	\Receiveddate{zzz} \Reviseddate{zzz} \Accepteddate{zzz} 
	
	\begin{abstract}
		Abstract: Survival analysis is a fundamental area of focus in biomedical research, particularly in the context of personalized medicine. This prominence is due to the increasing prevalence of large and high-dimensional datasets, such as omics and medical image data. However, the literature on non-linear regression algorithms and variable selection techniques for interval-censoring is either limited or non-existent, particularly in the context of neural networks. Our objective is to introduce a novel predictive framework tailored for interval-censored regression tasks, rooted in Accelerated Failure Time (AFT) models. Our strategy comprises two key components: i) a variable selection phase leveraging recent advances on sparse neural network architectures, ii) a regression model targeting prediction of the interval-censored response. To assess the performance of our novel algorithm, we conducted a comprehensive evaluation through both numerical experiments and real-world applications that encompass scenarios related to diabetes and physical activity. Our results outperform traditional AFT algorithms, particularly in scenarios featuring non-linear relationships.
	\end{abstract}

	%% maketitle must follow the abstract.
	\maketitle                   % Produces the title.
	
	%% If there is not enough space inside the running head
	%% for all authors including the title you may provide
	%% the leftmark in one of the following three forms:
	
	%% \renewcommand{\leftmark}
	%% {First Author: A Short Title}
	
	%% \renewcommand{\leftmark}
	%% {First Author and Second Author: A Short Title}
	
	%% \renewcommand{\leftmark}
	%% {First Author et al.: A Short Title}
	
	%% \tableofcontents  % Produces the table of contents.

	\section{Introduction}
	In recent decades, interval-censored data modeling has been a subject of intense research from methodological, theoretical, and applied perspectives \citep{van1993hellinger,huang,bio,kosorok}. Interval-censoring arises from data collection mechanisms that prevent the target event of a system from being observed directly because it is only known to lie within an observed interval. The boundary elements of these intervals belong to a sequence of examination times or check-ups. This type of data is commonly encountered in various fields including medicine, engineering, and social sciences; requiring specialized statistical methods.  
	
	Recently, the first machine learning contributions appeared in the literature, such as specific smoothing random forest algorithms \citep{kosorok}, a more classical tree ensemble random forest \citep{10.1093/biostatistics/kxz025}, gradient boosting \citep{ensembles}, or the first methodology tackling interval-censoring based on kernel machines \citep{travis2021kernel}. Notably, deep learning algorithms are scarce in this setting \citep{sun2023neural} and inexistent in the high-dimensional case, despite the recent progress survival analysis has undergone during the last years with right-censored outcomes \citep{zhong2021deep, lee2018deephit, steingrimsson2020deep,rindt2022survival,jia2022deep}.

	Although deep learning models hold potential advantages to handle high-dimensional data and non-linear dependencies, extending neural network models to an interval-censoring context- or, more generally, to survival analysis- brings several methodological hurdles. Traditionally, models for regression of interval-censored outcomes are trained by solving complex second-order optimization problems \citep{splines}. However, first-order methods only train neural networks on strongly non-convex problems \citep{annals}. Furthermore, many survival analysis models have a semiparametric nature, an area where deep learning literature is scarce in developing both efficient estimators and specific optimization techniques.

	The goal of this paper is to propose a modelling framework based on neural networks centered around AFT targeting interval-censored outcomes. The new methodology includes:
	\begin{enumerate}
		\item A sparse neural network architecture for variable selection with interval-censored responses based on \citep{lassonet}.
		\item A flexible predictive algorithm for interval-censored outcomes.
	\end{enumerate}
	
	From a technical perspective, AFT can be viewed as the natural generalization of linear regression to the field of survival analysis aiming to model conditional distributions  {between} two random variables $Z\in \mathbb{R}^{d}$ and $T\in \mathbb{R}$. It has a clear physics motivation and  {is widely used in practice} \citep{wei1992accelerated}. 
	
	Two main lines of research have been developed in the literature. The first involves specifying a parametric law for the noise that characterizes the conditional distribution of the scale and location model \citep{bagdonavicius2001accelerated}. In the second, the noise distribution is not specified, generally being estimated with nonparametric methods \citep{https://doi.org/10.1002/sim.8786}. However, this approach- popular in the field of right-censoring- poses challenges in potential extensions to the interval-censored case due to identifiability issues and numerical estimation problems. Therefore, during our simulated experiments and case studies, we will assume a parametric noise structure to obtain stable optimization algorithms.
	
	A critical step in our framework, especially in the variable selection case, is to consider \citep{meinshausen2010stability}'s stability selection. Stability selection is an alternative strategy for selecting variables in noisy environments, as is the case of our context. Random weight initialization in deep learning computational libraries helps estimating stability paths, contributing to the randomness arising from subsampling. Traditional approaches to selecting structure, such as cross-validation, are difficult to carry out because the initial weights' randomness leads to local minima, which generate discordant solutions for each fold and each structure, making it challenging to assess model stability.
	
	\subsection{The  {difficulty} of statistical inference with interval-censored data}
	
	The non-parametric estimation of survival functions with interval-censored data is based on \citep{turnbull1976empirical}'s
	algorithms, derivable with a self-consistency argument \citep{efron1967two} which he proved to be also the NPMLE. Significant issues pointed out in the literature are its lack of uniqueness if the log-likelihood is not strictly concave and the fact that consistency results are usually available upon assuming a fixed number of inspection times \citep{gentleman}. \citep{Maathuis2008-xg} deduce that the NPMLE for the joint distribution function of an interval-censored survival time and a continuous mark variable is not consistent in general. What is more, the self-consistent approach may not be consistent even when the inspection times only take on finitely many values, see counterexample in \citep{yu2000consistency}. Moreover, traditional methodology in the corpus of computational statistics such as the bootstrap is known to be inconsistent when using the NPMLE as the reference measure with interval-censoring \cite{SEN2015121}. Due to these challenges, practitioners often resort to parametric and semi-parametric approaches for modeling interval-censored data that are numerically more stable and yield consistent solutions from both optimization and statistical perspectives. In this paper to introduce more flexibility in the semiparametric modeling, we propose Accelerated Failure Time (AFT) predictive framework using a neural network models.

	%In addition, there exist many examples of MLE even in the parametric setting, see, \citep{bahadur1958examples}. This prompted further research and the development of alternative methods in the field of survival analysis to improve the reliability of statistical inference under interval-censoring. For example, \citep{lin97} claim that the naive approach of averaging on intervals can be misleading and they propose an intricate solution that involves operating with the Kaplan-Meier estimator. 

	\subsection{Summary of contributions}

	Our manuscript introduces a novel framework in the field of interval-censored data. The main contributions are discussed below:
	
	\begin{enumerate}
		\item We present the first variable selection algorithm in interval-censoring based on neural networks, leveraging advances in LassoNet-like architectures \citep{lassonet}. It is specifically tailored to select the most relevant covariates regarding the  {conditional survival function} of a time-to-event response. Unlike previous approaches in the literature that focused on variable relevance in terms of conditional expectations, our approach concentrates on modeling the conditional survival function in terms of AFT parametric models \citep{wei1992accelerated}.
		
		\item Variable selection becomes a challenge when estimators depend on numerous hyperparameters and difficult underlying optimization problems may lead to numerical instability, as is the case of neural networks. Similar problems occurred in a less severe way in traditional penalized regression models, as is the case of Lasso \citep{originalasso}. Practitioners have used solutions such as stability selection, which possess guarantees on the expected number of false positives under general conditions and from a model-free perspective. Here, we use stability selection  \citep{meinshausen2010stability} for the first time in the setting of neural networks and empirically validate its applicability and effectiveness.
		\item We introduce, from a predictive perspective, a non-sparse architecture that can be stably trained with first-order optimization methods and outperforms traditional AFT models under non-linear ground dependencies. 
		\item We deduce the analytical expression of a new family of curves that serve to guide practitioners graphically in terms of false discovery rate, see Figure \ref{wrap}.
		\item We illustrate the potential of our methodology through relevant biomedical applications. Thanks to the interpretability of AFT models, we obtain new findings in our application studies. 
		\item The software implementation of our method is freely available for public use and can be accessed through \url{https://github.com/meixide/spinet}.
		
	\end{enumerate}

	%Stability selection, on the other hand, is a robust variable selection method that accounts for this randomness by repeatedly subsampling the data and estimating the selection probability of each variable, taking a healthy advantage of random initialization.
	
	%Additionally, we propose to parametrize regression with a log-normal accelerated failure time (AFT) model. This model allows for a more flexible specification of the survival function, which can better capture the non-linear relationship between predictors and survival times. The log-normal AFT model assumes that the logarithm of the survival time follows a normal distribution with a mean that depends linearly on the predictors. This model has been shown to perform well in practice and has been used in various fields, including medicine and engineering.
	
	%The proposed approach involves splitting the sample into a training and a validation set. We use the stability selection method on the training set to select the most important variables and then fit a low-dimensional LassoNet model on the selected variables. We then evaluate the performance of the model on the validation set using appropriate metrics such as the concordance index. 

	%This paper proposes a novel approach for non-linear regression of time outcomes in the presence of interval-censored data using LassoNet with stability selection. Our approach provides a flexible and interpretable framework for analyzing interval-censored data, with important applications in personalized medicine and other fields.
	
	\subsection{Structure of the paper}
	The structure of the manuscript is detailed below. In Section 2, we introduce the general form of AFT models for both variable selection and prediction. We also discuss the interval-censored likelihood relevant to our problem and present its optimization scheme. Section 3 presents the results of a simulation study that demonstrates the advantages of our approach followed by a relevant case study related to kidney disorders from the T1D Exchange, a database containing important information on progression of individuals with type 1 diabetes mellitus in the United States.  In Section 4, we discuss the methods used, the results obtained, and future research questions raised by our approach. The Supplementary Material contains another relevant application related to the National Health and Nutrition Examination Survey (NHANES), a program of studies designed to assess the health and nutritional status of adults and children in the United States.

	\section{Methods}
	\label{sec1}

	Let $T \in [0,+\infty)$ be a random time-to-event response variable and $Z \in \mathbb{R}^d$ a random vector of covariates. We denote by $1-G(t\mid z)=P(T> t | Z=z)$ the  {conditional survival function}: the probability of event occurrence before $t \in [0,+\infty)$ given $z\in \mathbb{R}^d$.  
	In this paper, we restrict ourselves to AFT models, which can be expressed as $G(t\mid z)=F(\phi( t)-r(z))$, where $F$ is a noise distribution to be specified, $\phi$ is an appropriate monotone function and $r$ is a regression function approximated by elements belonging to a certain space that we will define subsequently. 
	\subsection{``Case-2'' interval-censoring}
	Interval-censoring is an ubiquitous data structure in medical studies, so it has to be faced by predictive models of disease developing. The methodology presented in this paper is specifically tailored for ``case 2'' interval censoring, which can be used to model more complex schemes. This is because, computationally speaking, the general interval-censoring scheme can be effectively reduced to ``case 2'' interval-censoring, which makes this approach a practical and powerful tool for handling a wide range of scenarios. We briefly describe three types of interval-censoring mechanisms following \citep{sun2006statistical}.
	
	In current status data- also known as ``case 1'' interval-censoring- the only knowledge about the failure is whether it has occurred or not before an observed checkup. “Case k” interval censoring, $k>1$, arises when there are $k$ examination times
	per subject. As a particularization with $k=2$, in ``case 2'', we can only determine that the true failure has occurred either within some random time interval, before the left border of the time interval, or after the right end point of the interval. We insist in that it is not possible to precisely determine the exact time at which the event occurred, but rather we can only observe its possible range of values. Last generalization consists on letting the number of observation instants to vary across individuals. Note that it is convenient to reduce this type of interval-censoring to ``case 2'' by considering whether the failure occurred before the first examination time, after the last examination time, or between two checkpoints. As we mentioned, this transformation locates ``case 2'' interval-censoring as a fundamental setting.

	Let $T_i$ be the $i$th patient's unobservable failure time for $i=1, \ldots, n$. In the framework of interval-censored data $T_i$ will never be observed.
	In ``case 2'' interval-censoring scheme, data can be expressed as
	$$\left\{\left(Z_{i},U_{i}, V_{i}, \delta_{1i}= 1_{\left[T_{i} \leq U_{i}\right]}, \delta_{2i}=1_{\left[U_{i}<T_{i} \leq V_{i}\right]}\right)\right\}_{ i=1}^n$$
	where $Z_i \in \mathbb{R}^d$ is the vector of observed covariates and $U_{i}\leq V_{i}$ a.s. are two independent random checkup time points for individual $i$. 
	Now, assuming as in \cite{wellner} that
	
	\begin{enumerate}
		\item the (unobservable) failure time is independent of the examination
		times given the covariates
		\item the joint distribution of the examination times and the covariates
		are independent of the parameters of interest
	\end{enumerate} 
	
	\noindent we can write the interval-censored log-likelihood function:

	\begin{align*}
		l_n(\sigma,r) = \sum_{i=1}^n &  \left\{
		\delta_{1i} \log F\left(\frac{\log U_i-r(Z_i)}{\sigma}\right)\right.
		\\+ & \delta_{2i} \log \left[F\left(\frac{\log V_i-r(Z_i)}{\sigma}\right)-F\left(\frac{\log U_i-r(Z_i)}{\sigma}\right)\right]\\ + & \left.\delta_{3i} \log \left[1-F\left(\frac{\log V_i-r(Z_i)}{\sigma}\right)\right]\right\}
	\end{align*}
	
	where $\sigma$ is a scale parameter; $\delta_{1i}$, $\delta_{2i}$, and $\delta_{3i}=1-\delta_{1 i}-\delta_{2 i}$ are indicator variables taking the value 1 if the $i$th observation is left, interval or right censored, respectively. 
	In instances where exact survival times, denoted as $T_i$, are observed, their contribution to the log-likelihood can be properly expressed as the evaluation of the standard normal probability density function (pdf) at $\frac{\log T_i - r(Z_i)}{\sigma}$.

	\subsection{Variable selection}
	\label{data}

	\subsubsection{Sparse neural networks}
\begin{center}
		\begin{figure}[h]
			
			\caption{Example diagram of a feedforward neural network with an skip layer. The architecture used in the application to NHANES data would be the same, adding an extra layer with the same number of hidden units (10). The total number of parameters, in that case, is $604$. Bias terms are omitted in the drawing.}\label{arch}
			\centering
			\includegraphics[width=0.5\textwidth]{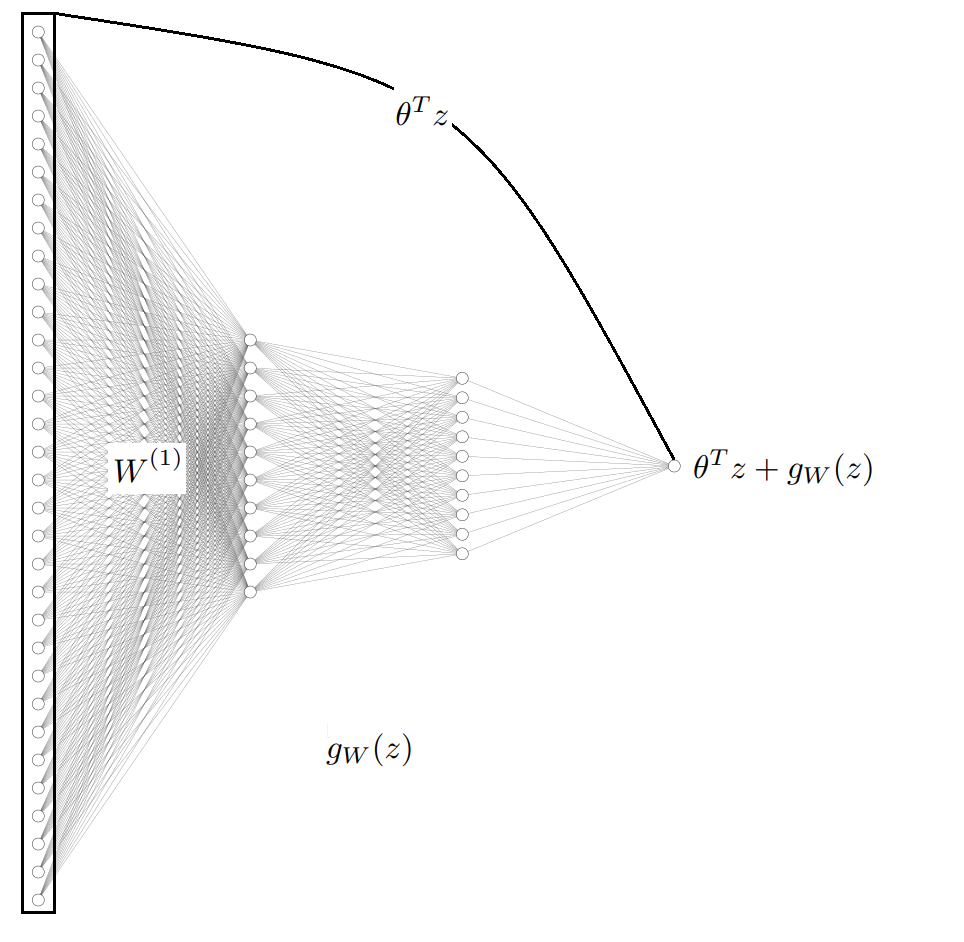}
		\end{figure}
	\end{center}

	Additional architectural constraints are needed to produce sparse estimators in the context of neural networks. In many applications, such as those in medical disciplines, the interpretability of parameter estimates plays a crucial role. Despite their prediction accuracy and generalization ability, deep learning models are built upon relatively large sets of parameters whose values are not informative for practitioners. 
	With sights on solving this problem, our methodology is inspired by the new neural network sparsity paradigm introduced in \citep{lassonet}. We depict its architecture in Fig. \ref{arch}

	The regression function involved in the AFT model specified in subsection \ref{data} will be approximated by elements 
	$r_{\theta, W} \in \mathcal{R}$,
	being $\mathcal{R}$ the space of deep residual blocks 
	\begin{align*}
		\mathcal{R}=\{&r_{\theta, W}: z \in \mathbb{R}^d \mapsto \theta^{T} z+g_{W}\left(z\right),\\&\theta \in \mathbb{R}^d,\quad g_{W} \text{ feed-forward NN with weights }W \}
	\end{align*}

	Recovering the identity $r_{\theta, W}\left( z\right) = \theta^{T} z+g_{W}\left(z\right)$ we can simply note that $g_{W}\left(z\right)=r_{\theta, W}\left( z\right) - \theta^{T} z$. This means that the deep learning contribution to the AFT model accounts for what linearity cannot explain. This is the idea behind the working principle of Residual Networks, which have been proven to be universal approximators \citep{nips}.

	We outline in Algorithm 1 a new optimization routine inspired by LassoNet's \verb|HIER-PROX| \citep{lassonet}. The whole optimisation problem is defined therein as
	$$
	\begin{aligned}
		&\underset{\sigma,\theta, W}{\operatorname{minimize}} \quad L(\sigma,\theta, W):=-l_n(\sigma,\theta, W)+\lambda\|\theta\|_{1} \\
		&\text { subject to }\left\|W_{j}^{1}\right\|_{\infty} \leq M\left|\theta_{j}\right|, j=1, \ldots, d,
	\end{aligned}
	$$
	where $W_{j}^{1}$ denotes the outward weights for feature $j$ in the first hidden layer. The objective is minimised via the technique of hierarchical proximal optimisation. 
	
	The proposed architecture integrates both linear transformations and a perceptron-based model acting on the input data. This dual-component structure offers a flexible way to capture different sparse patterns in data thanks to the constrained optimization problem above. However, it is important to note that the architecture introduces non-identifiability: there is no control on the signal proportion absorbed by each part, making it challenging to disentangle the unique contributions of each. While this does not harm the predictive power of the model, it offers an avenue for future research on interpretability and model decomposition.
	
	Because of the $\ell^1$-penalty placed on the objective function, the linear contribution of many input covariates is being shrunk to zero. The constraint imposes that the first hidden layer weights associated with such covariates are also being shrunk to zero. This fact has true regularization implications, as the non-linear effects of covariates that are screened by the linear Lasso scheme vanish all the way down forward propagation throughout the hidden layers. We provide an adaptation that makes compatible the optimization of the interval-censored likelihood with deep learning automatic differentiation computational libraries such as \texttt{torch.autograd} \citep{pytorch}\footnote{\url{https://pytorch.org/tutorials/beginner/blitz/autograd_tutorial.html}}. The high level algorithm is given in Algorithm \ref{alg:cap} below.

	\begin{algorithm}
		\caption{Optimization routine derived from \citep{lassonet}}\label{alg:cap}
		\begin{algorithmic}[1]
			\State Initialize $\theta^{(0)}$, $W^{(0)}$, $\sigma^{(0)}$, number of epochs $B$, hierarchy multiplier $M$, learning rate $\alpha$, number of iterations $K$
			
			\For{$k \in\{1 \ldots K\}$}
			
			\State Compute gradient of loss function w.r.t $\left(\sigma,\theta, W\right)$ using back-propagation \State Update $\sigma \leftarrow \sigma-\alpha \nabla_{\sigma} L$ , $\theta \leftarrow \theta-\alpha \nabla_{\theta} L$ and $W \leftarrow W-\alpha \nabla_{W} L$
			
			\State Update $\left(\theta, W^1\right) \leftarrow$ \verb|HIER-PROX| $\left(\theta, W^1, \alpha \lambda, M\right)$

			\EndFor
			
		\end{algorithmic}
	\end{algorithm}

	\subsubsection{Stability selection}
	
	%Stability selection is a method for estimating discrete structures or variable selection, which has become increasingly important due to the growth of high-dimensional data in many scientific disciplines. The method addresses the problem of proper regularisation, which is crucial for obtaining the right-sized structure or model. It does this by using a subsampling approach to determine the amount of regularisation needed to conservatively control for multiple testing. This approach not only improves the selection of variables but also yields a new structure estimation scheme. The method has been shown to have promising properties, particularly for high-dimensional problems, and has been used for finite sample familywise error control and improved structure estimation or selection.
	
	%In our case, we are also taking advantage of the non-deterministic selection process of the variables by the neural network. The randomness of the variable selection process is driven by the initialization of the weights of the network, which allows for additional improvements in the selection process. 

	Stability selection \citep{meinshausen2010stability} is a powerful method for estimating discrete structures and performing variable selection, especially in high-dimensional data contexts. The main challenge in such scenarios is finding the right amount of regularization to achieve an optimal model size. To address this issue, stability selection uses a subsampling strategy to assess variable selection stability across different data subsets, enabling control over the expected number of falsely selected variables. This method improves the accuracy of variable selection and provides a new structure estimation scheme. 
	
	In our case, we are leveraging the non-deterministic variable selection process of neural networks, which in other grid-search-based methods would constitute a computationally infeasible task. The initialization of the network's weights introduces a beneficial level of randomness that can be more effectively utilized in our setting. 
	
	One critical advantage of stability selection is its ability to provide a guarantee on the expected number of false selections, under certain assumptions \citep{meinshausen2010stability}. Stability selection involves repeatedly subsampling a dataset without replacement, each time applying a variable selection method under varying levels of regularization. For each regularization level, it counts how frequently each variable is selected across these subsamples. By setting a threshold on this frequency, one can ensure that the expected number of falsely selected variables is bounded, thus enhancing the reliability of the variable selection process. This method provides a principled way to choose the regularization parameter, thereby controlling the rate of false discoveries in the model.
	
	Assuming that the distribution of selected variables is exchangeable and the original procedure is no worse than random guessing, the expected number of falsely selected variables can be bounded: 
	$$E(V) \leq \frac{1}{2 \pi_{t h r}-1} \frac{q_{\Lambda}^2}{d},$$
	
	where $q_{\Lambda}$ is the average number of selected variables across the regularisation parameter set $\Lambda$, $\pi_{thr}$ is the cutoff parameter and $d$ is the total number of covariates as denoted throughout the rest of the paper.
	
	Stability selection is a helpful tool for variable selection in high-dimensional data analysis. Traditional methods, such as cross-validation, can be challenging to use for deep learning problems in high-dimensional settings because the risk curve, which shows the relationship between the model complexity and prediction error, can be too noisy and hard to minimize. In contrast, stability selection uses subsampling to control the selection of variables and provides a more stable and interpretable solution. 
	
	\subsection{Predictive model}
	Regression models serve as essential tools for understanding relationships among variables. They allow for the assessment of interaction terms, non-linear effects, and other complex relationships within the data. 
	
	In this operating mode of our proposed method, the regression function will be approximated by elements 
	$r_{W} \in \mathcal{P}$, 
	being $\mathcal{P}$ the space of feedforward neural networks with no skip layer 
	\begin{align*}
		\mathcal{P}=\{r_{W}: z \in \mathbb{R}^d \mapsto g_{W}\left(z\right),g_{W} \text{ feed-forward NN with weights }W \}. 
	\end{align*}

	Unlike variable selection, which primarily focuses on reducing dimensionality and identifying significant predictors, predictive modeling extends our analytical capabilities.

	\subsection{Sample-splitting}
	
	Our multi-stage regression method involves three steps, following a similar lens to \citep{wasshigh}. A generic way for performing high-dimensional regression consists on splitting the sample with indices $\{1, \ldots, n\}$ into a partition $ {D}_1,  {D}_2 \text { and }  {D}_3$. 
	\begin{enumerate}
		\item Stage I. Use $ {D}_1$ to find $\hat{S}\left(D_1\right) \subset\{1, \ldots, p\}$ using stability selection. 
		\item Stage II. Use the subsample $ {D}_2$ and the selected variables $\hat{S}\left(D_1\right)$ to find the low dimensional estimate $r_{ \hat W}$ for model $\hat{S}\left(D_1\right)$. 
		\item Stage III. Use $ {D}_3$ and $r_{\hat W}$ to test the performance of the model on unseen data using C-index \citep{harrellc}. 
		
	\end{enumerate}
	
	An scheme for our proposal is the following:
	$$\underbrace{\text {sample} \stackrel{\text { stage I }}{\longrightarrow} \hat{S}\left(D_1\right) \stackrel{\text { stage II }}{\longrightarrow}}_{\text {train }} r_{\hat W} \underbrace{\stackrel{\text { stage III }}{\longrightarrow}\text {C-index}}_{\text {test}} .$$
	We make explicit the third step involving evaluation on a fresh dataset, in contrast with classical splitting structures \citep{hdi}.

	\section{Simulated experiments}
	\label{sec2}

	In this paper, we have opted for the most straightforward configuration to maintain clarity and facilitate illustration.
	We fix $F$ to be the standard normal cdf and $\phi(t)=\log(t)$. This leads to the AFT $\log T=r_{\theta, W}(Z)+\varepsilon, \quad r_{\theta,W} \in  \mathcal{R}$
	with $\varepsilon \sim \mathcal{N}(0,\sigma^2)$. The parameter space is 
	$$\Gamma=\left\{ \left(\sigma,\theta,W \right) \in (0,+\infty) \times \mathbb{R}^d \times \mathbb{R}^N \right\}$$ 
	where $N$ is the total number of weights involved in the layers of the deep residual block.  
	
	While more complex settings could offer additional nuances, the aim here is to provide a transparent and easily interpretable analysis, thereby serving as a foundation for future studies that may wish to explore these complexities in greater depth.
	
	%\subsection{Simulation studies}
	
	%This section presents studies to assess the performance of our proposed method.
	
	\subsection{Generation of interval-censored targets}
	
	We describe a method to simulate interval-censored data, resembling the scenarios encountered in medical studies with periodic checkups. A fixed number of checkup times (20) and the probability of attending the next checkup (0.5) are set. 
	
	The logarithm of latent event times is generated based on the mean vector (see further details below) plus random normal zero-mean noise. Each patient's checkup times are then randomly generated within a log-uniform distribution. 
	
	Next, an algorithm determines whether each checkup was attended before or after the latent event time. Based on this, it detects the nearest checkup time before and after the event. The delta indicators save whether the event time is left-censored, interval-censored  or right-censored.
	
	This synthetic data generation method closely mimics the data collection process in medical studies, where patients undergo periodic checkups, and the exact event time (like onset of a disease) is unknown but is bounded by these checkup times. This approach effectively creates interval-censored data, essential for testing and validating statistical models designed for such data structures.
	
	\subsection{Description of the simulation setups and results}
	
	We describe the structure of two simulated regression scenarios below. Example 1 generates the regression function according to the following model:
	$$\log T=X_1+X_1^2-0.5\left(X_2+X_2^2\right)+0.5\left(X_3+X_3^2\right)+1 +\varepsilon$$
	\noindent where $X_1, X_2,\ldots, X_{100}$ are the predictor variables with $X_i \sim \mathcal{N}(0,1)$ for $i = 1, 2, 3$. The error term is $\varepsilon \sim \mathcal{N}(0,\sigma^2)$, where $\sigma = 0.5$. The covariance matrix for the predictor variables is the identity. 
	
	For Example 2, let $\beta_0=-2$, $\beta_1=0.5$, $\beta_2=-0.7$, $\beta_3=0.8$, $\beta_4=0.6$, $\beta_5=-0.3$, and $\sigma=0.5$. We also specify the covariance matrix $\Sigma$ for the predictor variables as $\Sigma_{i, j}=(-1)^{i-j} \rho^{|i-j|}$ where $\rho=0.3$ and $i,j=1,\dots,100$. We then generate the predictor variables $X_1, X_2, \ldots,X_{100}$ by drawing a sample of size $n$ from a multivariate normal distribution with mean vector ${0}$ and covariance matrix $\Sigma$. Finally, the true survival times $T$ are calculated as a function of the predictor variables:
	\begin{align*}
		\log T&=\beta_0+\beta_1 X_1+\beta_2 X_2+\beta_3 X_3&\\&+\beta_4 X_1 X_3+\beta_5 \log \left(X_2^2+1\right)&+\varepsilon
	\end{align*}
	
	\noindent where $\varepsilon$ is a random error term drawn from a normal distribution with mean $0$ and standard deviation $\sigma$. 
	
\begin{figure}
		\centering
		\includegraphics[width=0.7\linewidth]{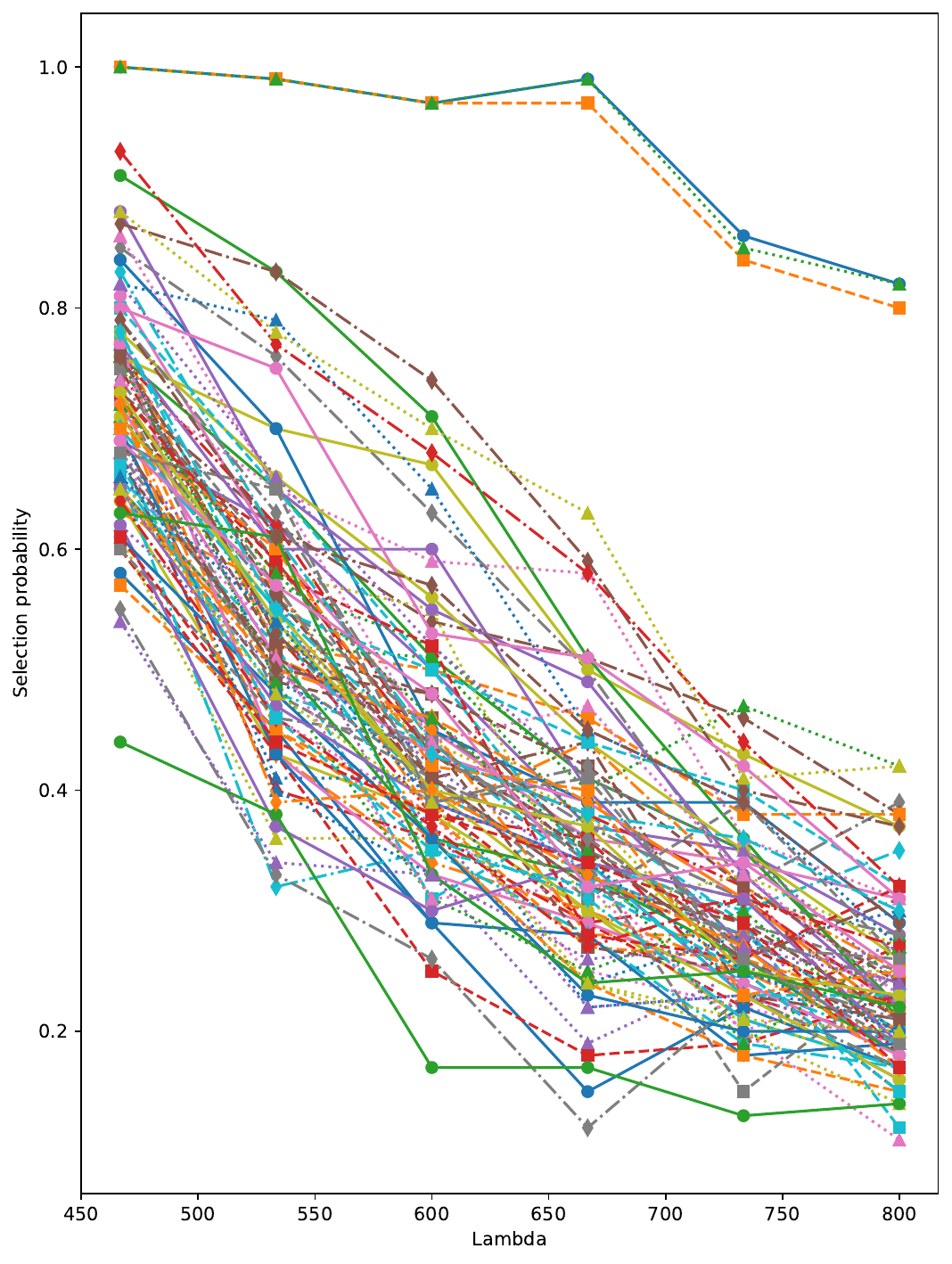}
		\caption{The stability selection technique applied to synthetic data: example 1 with $n=1000$.}
		\label{fig:sub1}
	\end{figure}

	To  {assess the variable selection performance} of our proposed method in the two datasets described above, we generated 500 simulated replicas for each scenario, varying the sample size between $n=500,1000,2000$. Stability selection was then run across $100$ subsamples and $50$ values of $\lambda$ to estimate the stability paths of each simulation. Our results demonstrate that the variables were consistently and accurately selected, indicating the robustness and reliability of our method. See Figure \ref{fig:sub1} for a graphical illustration.

	In the predictive regression experiments, we evaluate the performance of our method (Table \ref{tab:example1}) against \verb|LogNormalAFTFitter| from the \verb|lifelines|\footnote{\url{https://lifelines.readthedocs.io/}} \citep{davidson} package (Table \ref{table:tabla2}), using the same simulated datasets after manually removing irrelevant variables from the predictors. We measure the predictive accuracy in survival analysis using the concordance index (C-index, \cite{harrellc}), which is computed over a fresh test sample of the same size and generated by the same mechanism as the original dataset. C-index is computed as 
	
	$$C-index=\frac{\sum_{i \neq j} 1\left\{\eta_i<\eta_j\right\} 1\left\{t_i>t_j\right\} (1-\delta_{3j})}{\sum_{i \neq j} 1\left\{t_i>t_j\right\} (1-\delta_{3j})},$$
	
	where $\eta_k$ is minus the predicted conditional mean of log-survival time given by the fitted model;  $t_k$ is the mid-point of interval $k$ in case of left or interval censoring and left border in case of right-censoring. 
	
	Additionally, we use the coefficient of determination $R^2$, which measures the strength of the relationship between the predicted values of the AFT regression surface and the original noiseless logarithm of underlying times. To assess performance across simulated datasets, we average the metrics across simulation runs. Figure \ref{fig:sub1} demonstrates the effective discrimination of relevant and irrelevant variables in Example 1 using stability selection.
	
	The key findings from the simulation section are summarized as follows:
	
	\begin{itemize}
		\item In Example 1, our method outperforms the \texttt{LogNormalAFTFitter} when clear nonlinearity in statistical association is present. Our model exhibits a 40\% higher predictive power in terms of performance measures. Additionally, as the sample size increases, our model converges to the $\sigma$ parameter, while the competitor deviates significantly from the true value due to model misspecification.
		
		\item In Example 2, when the data structure closely resembles linearity, both our proposal and the \texttt{LogNormalAFTFitter} yield comparable accuracy. Nonetheless, our model excels at capturing subtle nonlinearities, a capability beyond the reach of the competitor. Additionally, it successfully converges to the true value of the $\sigma$ parameter.

		\item In both examples, the variable selection algorithms yield no false positives with $n=2000$, and the true positives are correctly identified. This demonstrates that in these scenarios, the variables are detected accurately.

	\end{itemize}
	
	Overall, our proposal outperforms the competitor, even when the model is quasi-linear. This highlights the broader applicability of our approach in approximating more complex nonlinear regression functions and justifies the flexibility of our new framework. 
	
		\section{Practical considerations}\label{wrap}
	
	The stability selection procedure requires a careful implementation to ensure its effectiveness when particularized to our proposal. This section outlines the  framework that encapsulates the core algorithm, following the principles laid out in the original LassoNet \citep{lassonet} and stability selection \citep{meinshausen2010stability} papers, with adaptations to suit our specific context and to enhance interpretability of results. This implementation allows users to effectively explore the relationship between the proportion of selected variables ($q_\lambda/p$) and the selection frequency of each feature $j$ ($\pi^\lambda_j$), while also providing a graphical framework for controlling the expected number of falsely selected variables.
	
	Orientative limits for the lambda grid are set based on a preliminary whole path run (see Section 4.1 in \cite{lassonet}). For each subsampling draw, a distinct path is created. Next, we iterate through the selection states of all paths at each lambda value. During each iteration, the average selection probability for each variable is calculated across all models. Importantly, we ensure that the selection probability for each variable is monotonically decreasing as we progress through the lambda values. At each step, we take the element-wise minimum between the current selected and the new average selection probability. The outcome of this process is that the selection probability curves for each variable will be monotonically decreasing across the lambda values. Subsequently, we plot\footnote{This approach follows the guidance provided in the original stability selection paper, which states: \textit{``To do this, we need knowledge about $q_{\Lambda}$. This can be easily achieved by regularization of the selection procedure $\hat{S}=\hat{S}^q$ in terms of the number of selected variables $q$, i.e., the domain $\Lambda$ for the regularization parameter $\lambda$ determines the number $q$ of selected variables, i.e., $q=q(\Lambda)$. For example, with $l_1$-norm penalization as in expressions (2) or (4), the number $q$ is given by the variables which enter first in the regularization path when varying from a maximal value $\lambda_{\max }$ to some minimal value $\lambda_{\min }$. Mathematically, $\lambda_{\min }$ is such that $\left|\cup_{\lambda_{\max }\geq \lambda \geq \lambda_{\min } }\hat{S}^\lambda\right| \leq q$.''}} $(\pi^{\lambda}_{1}, \ldots, \pi^{\lambda}_{p})$ against $\frac{q_{\lambda}}{p}$.

	We recall the bound for the expected number of falsely selected variables under the assumptions in Theorem 1:
	
	$$E(V) \leq\frac{1}{2 \pi_{\mathrm{thr}}-1} \frac{q_{\Lambda}^2}{p}$$
	
	\noindent Dividing both sides by $p$, we obtain
	
	$$\frac{E(V)}{p} \leq \frac{1}{2 \pi_{\mathrm{thr}}-1} \frac{q_{\Lambda}^2}{p^2} =:F\left(\frac{q_{\lambda}}{p}\right)$$
	
\begin{figure}
		\centering
		\includegraphics[width=0.8\linewidth]{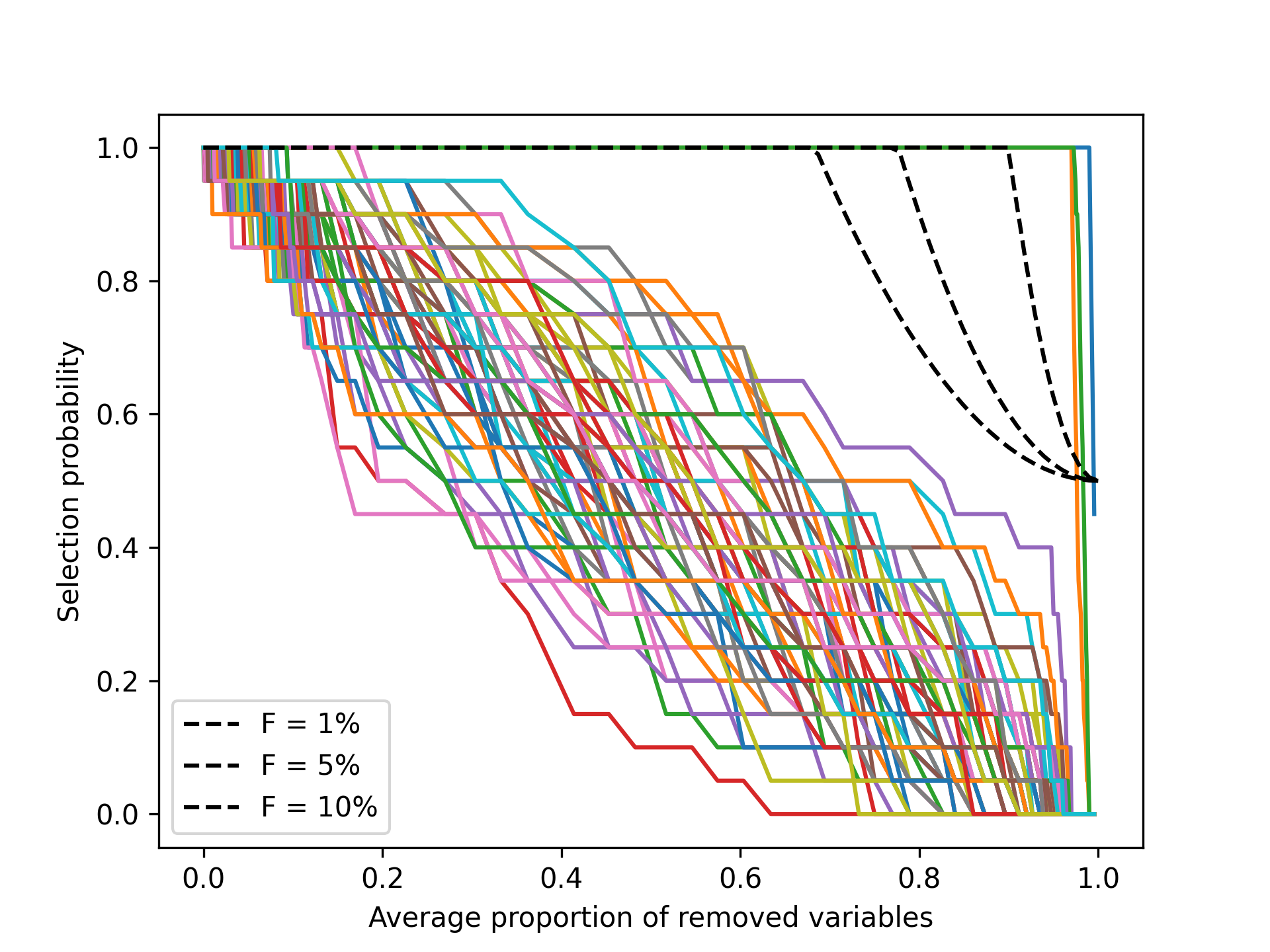}
		\caption{Graphical deliverable of Section \ref{wrap}. For a desirable expected proportion of removed variables $x$ and a false discovery rate  $F$ both fixed by the user, the variables eventually entering the model are the ones whose frequency curve $\pi_j$ surpasess the dotted black line given by $F$ at $x$}
		\label{fig:spinet}
	\end{figure}

	Let us define $x:=\frac{q_{\lambda}}{p}$. Then, for each bound $F$ on the expected proportion of falsely selected variables $\frac{E(V)}{p}$, we derive a family of curves indexed by $F$: $$\pi_{thr}(x)=\frac{1}{2}\left(\frac{x^2}{F} + 1\right).$$ Finally, we can add these curves to the $(\pi^{\lambda}_{1}, \ldots, \pi^{\lambda}_{p})$ against $\frac{q_{\lambda}}{p}$ plot to identify which variables would be selected upon fixing the false discovery rate upper bound $F$ we are willing to accept, see Figure \ref{fig:spinet}.

	\section{Application of the Method in the Type 1 Diabetes Exchange Registry Dataset}
	
	Survival analysis modeling is a powerful statistical approach used in medical research to understand the occurrence time-to-event clinical events, such as death and the development of  diseases as the case of Diabetes Mellitus. In the case of diabetes researchers, survival analysis plays an important role in the control, definition, and establishment of new therapeutic strategies for this heterogeneous metabolic syndrome. For example, the cutoff of the biomarker glycosylated hemoglobin [A1c] (a surrogate measure of the glucose values of the preceding $3$ months of glucose values) to diagnose diabetes Mellitus was derived from survival analysis models in terms from diabetes complications.
	
	From a technical perspective, diabetes disease is diagnosed and monitored with lab measures that are only collected between consecutive medical visits. Then, the different clinical events that are defined in diabetes clinical practice and by diabetes researchers lead to outcomes represented as interval-censored variables. To illustrate the importance of analyzing interval-censoring outcomes and to continue our discussion on the significance of these models in diabetes research, we proceed with a real example in this domain.
	
	In this study, our focus is on quantifying the progression of type 1 diabetes, particularly in terms of renal complications—a crucial criterion for considering the use of more intense and specific pharmacological treatments with therapeutic benefits, albeit accompanied by more secondary adverse effects.
	
	Here, the use of survival models to analyze the statistical associations between various risk factors and time-to-event outcomes serves multiple purposes in translational clinical research: i) managing type 1 diabetes effectively; ii) improving the metabolic glucose capacity of the patients; iii) minimizing mortality rates associated with type 1 diabetes; iv) reducing healthcare costs. Specifically, we measure the time until reaching a low albuminuria concentration, a critical surrogate marker for renal function specified in each medical lab test of the patients. The outcome exhibits an interval-censored structure because the clinical event happens between two consecutive medical examinations.
	
	For our analysis, we leverage data from the T1D Exchange, publicly available at \url{https://t1dexchange.org/}. The dataset includes information from 5971 patients, and the variables selected by our approach are summarized in Table \ref{table:selected}.
	
	Our primary goal is to create a risk score, represented as an Accelerated Failure Time (AFT) regression surface, for predicting the development of albuminuria concentration disorders in type 1 diabetes patients. Before showing AFT results, we start with a descriptive analysis that provide content to the sample analyzed.
	
	A descriptive analysis in the groups of patients that experienced the clinical event and those who did not reveals that the range of age involves patients with juvenile diabetes type 1 and all age ranges include elderly patients. For each example in the corresponding groups for age and BMI, the 25-75\% quantiles are 7.0 and 69.4, respectively, for individuals who experienced the event of interest. Conversely, for those who did not experience the event, these quantiles are 6.0 and 63.0. This trend aligns with the general expectation that older individuals are more prone to developing renal complications, suggesting age as a potential risk factor. Regarding BMI, the distribution appears relatively stable across both groups. Among those who experienced the event, the 2.5\% and 97.5\% quantiles for BMI are 15.5 and 38.0, respectively. In contrast, for individuals who remained at risk by the end of the study, these quantiles are 15.4 and 36.4. This similarity in the distribution of BMI across groups indicates that BMI might not undergo significant changes in relation to the observed event, suggesting a lesser or different role of BMI compared to age in the context of this study's event of interest.
	
\begin{figure}[h!]
		\centering
		\includegraphics[width=\linewidth]{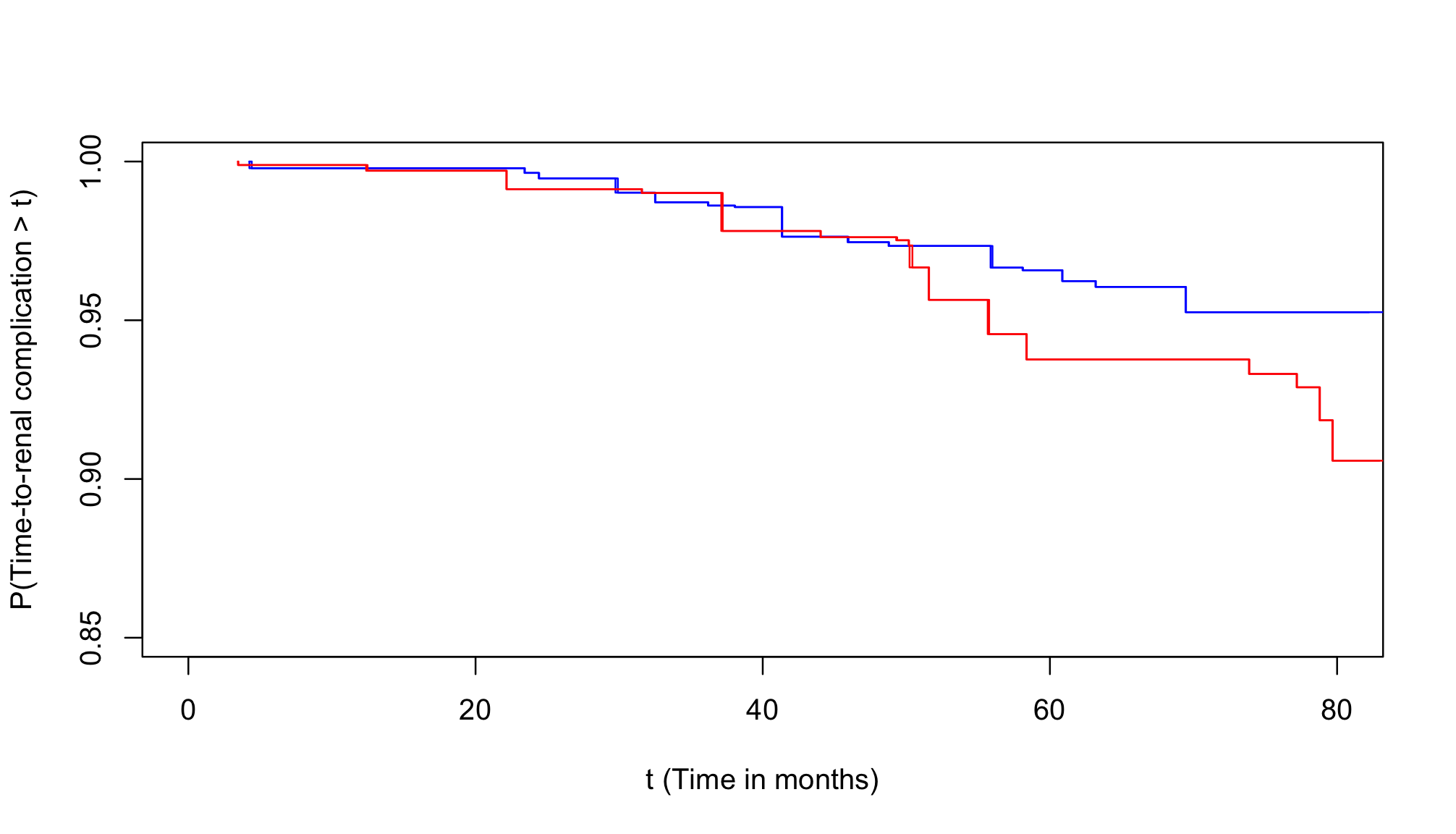}
		\caption{Marginal Turnbull estimators stratified by HbA1c values. Blue curve encompasses data satisfying HbA1c $<8$ and red involves observations such that HbA1c $\geq 8$.}
		\label{fig:turn1}
	\end{figure}

\begin{figure}[h!]
		\centering
		\includegraphics[width=\linewidth]{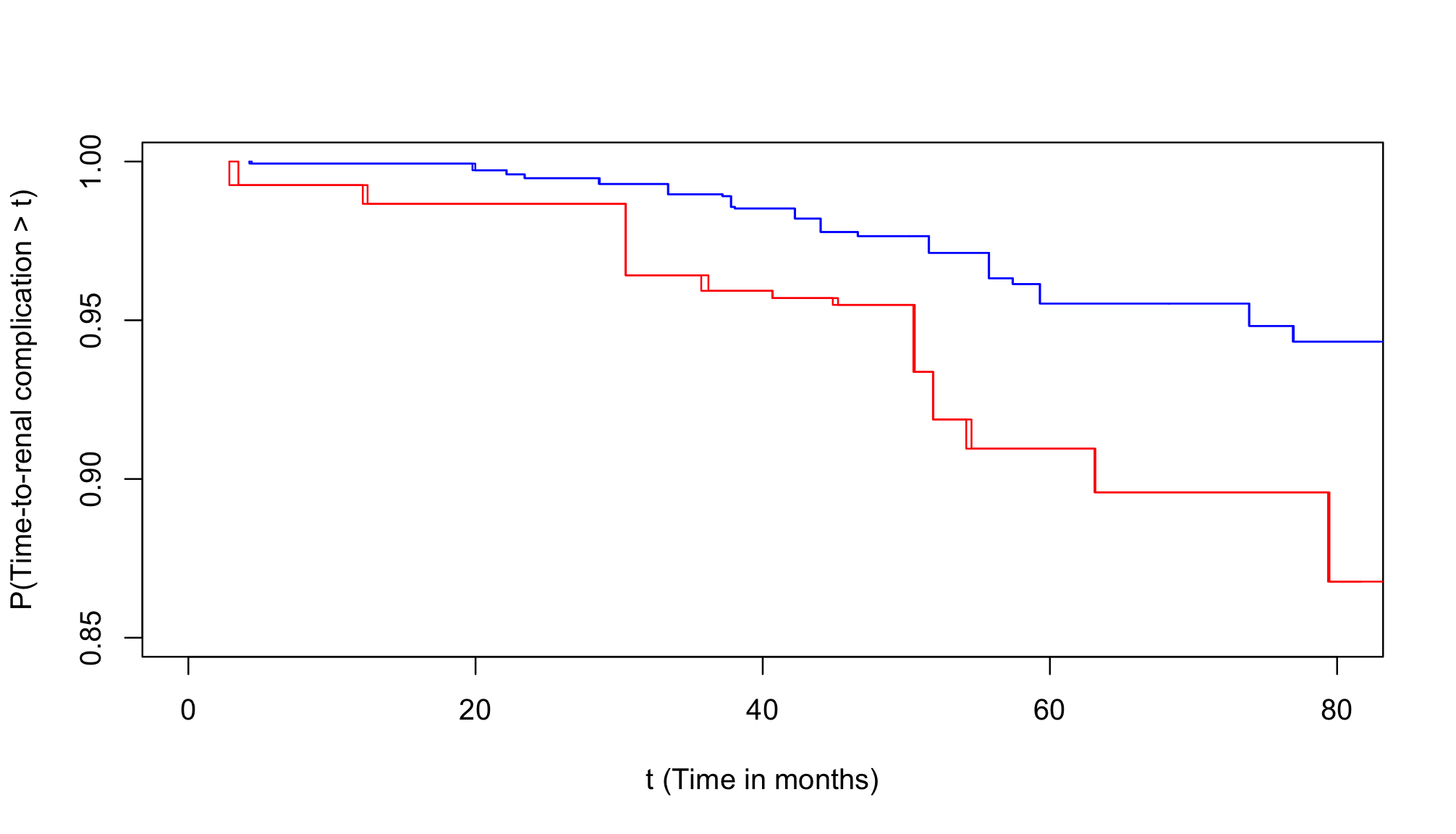}
		\caption{Marginal Turnbull estimators stratified by treatment indicator. Blue curve encompasses patients who did not take ACE inhibitors or AR blockers and red involves individuals who did.}
		\label{fig:turn2}
	\end{figure}

	To explore the temporal component when the clinical event occurred, Figures \ref{fig:turn1} and \ref{fig:turn2} present stratified Turnbull estimators categorized by A1c levels and treatment indicators, respectively. For A1C levels, we stratify the patients into well-controlled and poorly controlled diabetes patients. The marginal survival analysis indicates that fitting integrated regression models with several patient factors must be done to analyze the joint patient characteristics.
	
	After fitting the new neural network approach and comparing it with a classical AFT model with a log-normal distribution, our new models outperform traditional linear log-normal AFT models, achieving a concordance index of 0.65 compared to the traditional approach's 0.54. The concordance index was computed using a 20 percent left-fresh sample to assess the accuracy of both models. A summary of the selected variables can be found in Table \ref{table:selected}.
	
\begin{figure}
		\centering
		\includegraphics[width=0.5\linewidth]{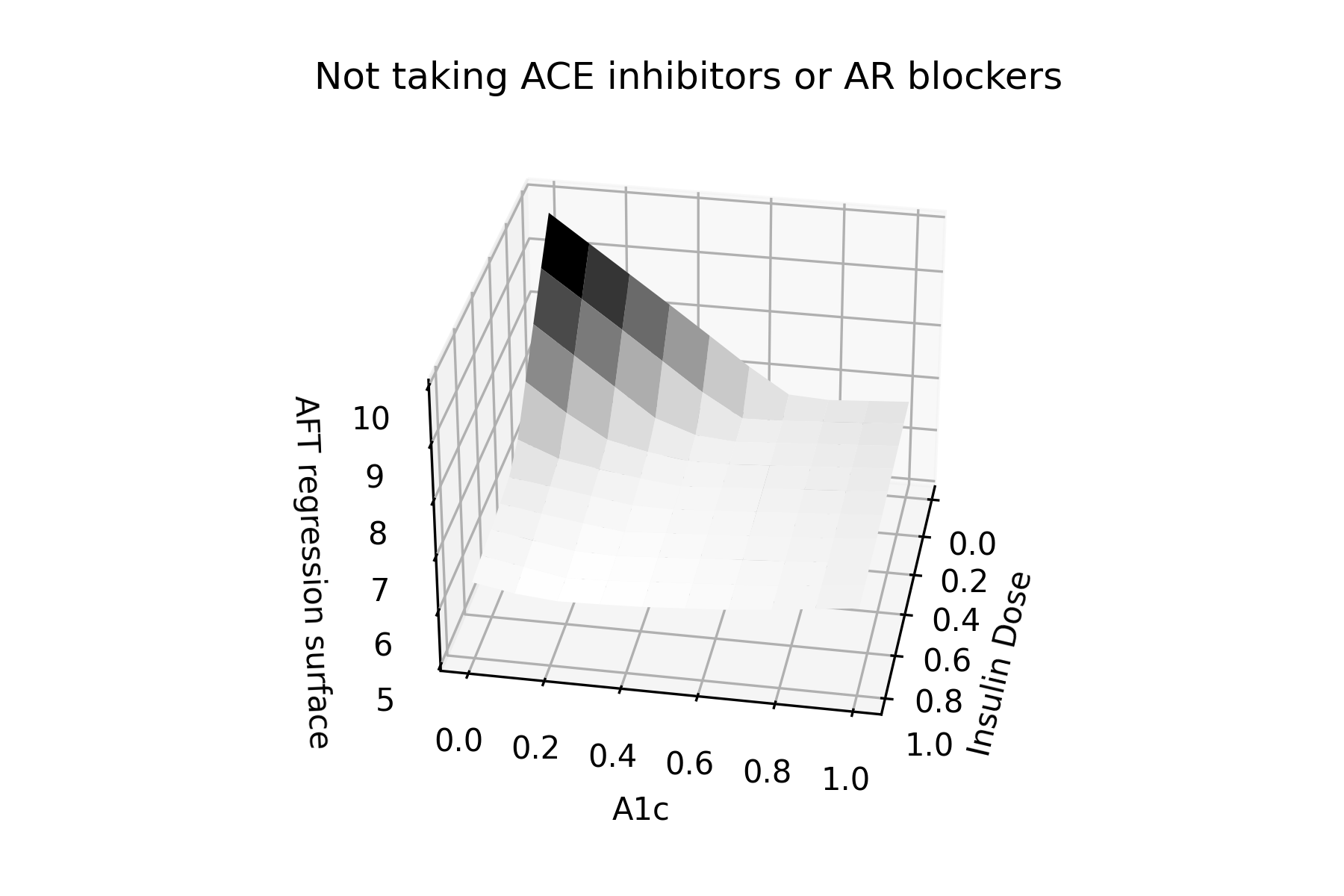}\includegraphics[width=0.5\linewidth]{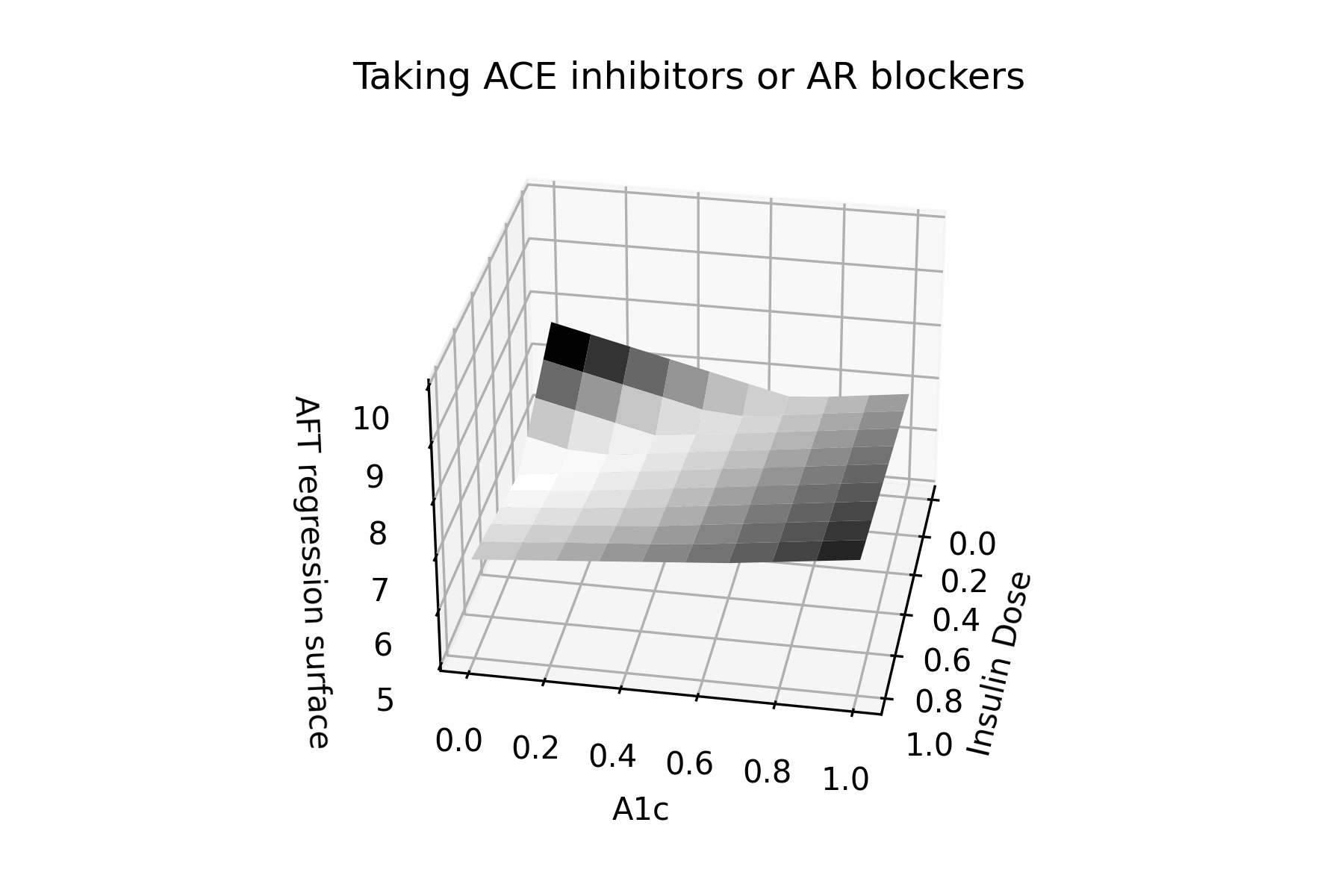}
		\label{fig:res}
		\caption{Predicted regression surface of our neural AFT model upon setting all the covariates to baseline value except from A1c and received units of insulin. Each plot corresponds to setting the treatment indicator to 0 or 1 in addition respectively, being the arms labeled in the title of the plots. Minimum and maximum values of these covariates are encoded in the plot as 0 and 1 respectively. This figure pictures the general idea that, if certain assumptions are met, neural networks can capture complex, non-linear interactions between features without explicitly modeling them. Logarithm of survival time as a function of these two variables is already highly non-linear per se, and when a third categorical variable comes into play we see that the regression surface is not modified by a simple shift. For the control arm, it is harder to reduce risk of albuminuria concentration disorder by providing insulin to patients with high values of A1c.}
		\label{inter}
	\end{figure}

	Our neural network learning-based approach offers preliminary valuable guidelines for delaying renal complications as indicated by disorders in albuminuria concentration. From a practical standpoint, our model identifies key variables such as A1c levels (a summary variable of the patients in the previous three months) and insulin dosage (the standard clinical treatment in this setting), providing a nuanced understanding of their impact on time-to-disorder. Figure \ref{fig:res} suggests that to achieve better outcomes, it is crucial to maintain minimal levels of both A1c and insulin dose. Additionally, we observe that angiotensin-converting enzyme (ACE) inhibitors and angiotensin receptor blockers (ARBs) have a modifying effect on the benefits of insulin administration.
	
	In cases of high A1c levels, insulin administration can delay time-to-pathology, and this beneficial effect is more significant in patients who take angiotensin-converting enzyme (ACE) inhibitors or angiotensin receptor blockers (ARBs). Patients with diabetes may require insulin therapy to help control their blood glucose levels. ACE inhibitors and ARBs are medications used to treat hypertension and heart failure, which block the action of angiotensin to reduce blood pressure and also have additional benefits such as reducing inflammation and improving insulin sensitivity. In general, in the realm of type 1 diabetes management, achieving optimal patient outcomes is a multifaceted challenge involving various pharmacological interventions and lifestyle modifications \citep{lewisne}, and our model provides new clinical knowledge about potential therapeutic interactions regarding A1c and insulin levels.
	
	In summary, our analysis reveals the following key findings: i) the predictive advantages of our model compared to Accelerated Failure Time (AFT) linear models in this application; ii) the sparsity of the underlying model, emphasizing the paramount importance of variable selection; iii) in the context of diabetes applications, the necessity to consider non-linear models due to the non-linear interactions illustrated in Figure~\ref{fig:res}; iv)  {the use of a larger pool of biomarkers and information from continuous glucose monitoring may be required to improve predictive capacity \citep{glucoriginal}}.
	
	% Start a two-column environmen
	\begin{table}[h!] 
		\small
		\centering
		\caption{Variable characteristics in the T1DX data.}
		\scalebox{0.8}{
			\begin{tabular}{|l|l|l|l|l|}
				\hline
				Variable & T1DX code & Description & Support & Selected \\
				\hline
				1 & ageConsent & Age at consent date (years) & Continuous & no \\
				2 & Pt\_Gender & Gender & 0: male, 1: female & no \\
				3 & BldPrSys & Systolic blood pressure & Continuous & no \\
				4 & BldPrDia & Diastolic blood pressure & Continuous & no \\
				5 & ACEARB & Taking ACE inhibitors or AR blockers & 0: no, 1: yes & yes \\
				6 & WeightKg & Participant weight in kilograms & Continuous & no \\
				7 & HeightCm & Height at exam (cm) & Continuous & no \\
				8 & diabDur & Duration of T1D (years) & Continuous & no \\
				9 & TotalDailyInsPerKg & Total daily insulin dose per kg & Continuous & yes \\
				10 & LDL & LDL value & Continuous & no \\
				11 & HDL & HDL value & Continuous & no \\
				12 & TotChol & Total cholesterol value & Continuous & no \\
				13 & Triglyc & Triglycerides value & Continuous & no \\
				14 & HbA1c & HbA1c & Continuous & yes \\
				15 & isletCellTrans & Participant had an islet cell transplant & 0: no, 1: yes & yes \\
				16 & bmi & BMI & Continuous & no \\
				17 & NumPumpBolusOrShortActUnk & Unknown number of boluses or insulin injections & 1: checked & yes \\
				\hline
		\end{tabular}}
		\label{table:selected}
	\end{table}
	
	%\onecolumn

	\section{Discussion}
	\label{sec4}

	This paper introduces a novel methodological contribution by presenting a new deep AFT model class for interval-censored targets. We introduce a novel variable selection strategy that leverages sparse neural networks, which is integrated in the regression algorithm. To the best of our knowledge, this is one of first methods combining the interval-censoring paradigm with deep learning algorithms. Given the prevalence of interval-censoring in personalized medicine applications, this model holds great relevance in this domain. The diverse results presented in this paper substantiate its value as an extension of classical AFT modelling while retaining interpretability advantages.
	
	Our work paves the way for new and exciting research directions in the field of interval-censoring. This includes addressing the noise distribution of AFT non-parametrically \citep{https://doi.org/10.1002/sim.8786}. Moreover, there is a growing need to provide measures of uncertainty for new predictions, particularly in medical science applications, where patient evolution inherently involves uncertainty. We propose extending conformal inference ideas to the interval-censoring setup, building upon existing approaches for right-censored data and split conformal inference methods \citep{izbicki}. Another intriguing possibility is to accommodate functional predictors such as medical images or wearable signals—a largely unexplored area with significant implications for digital medicine applications \citep{matabuena2023distributional, ghosal2023semiparametric}. For example, in our diabetes applications, the use of continuos glucose monitoring can improve the capacity prediction to detect long-term diabetes complications.

	Furthermore, establishing the significance of each selected variable represents another crucial research direction. Recently, post-selection methodologies for neural networks have emerged, and they can be applied to our model \citep{deeplink}.

	\section*{Appendix: NHANES physical activity data}
	
	The National Health and Nutrition Examination Survey (NHANES) is a comprehensive program conducted by the National Center for Health Statistics (NCHS), with the primary objective of collecting health and nutrition data from the United States population. In this study, we utilized a subset of $n=2977$ patients aged between 50 and 80 years, gathered from NHANES waves 2003-2006. This subset contains valuable data on physical activity patterns, measured with Actigraph accelerometry technology during 3-7 days with a time-resolution of one record per minute.
	
	From a public health and epidemiological perspective, establishing a statistical association between physical activity and mortality in the elderly population is of paramount importance. This analysis can serve as the foundation for promoting healthy habits across the population and designing targeted interventions to mitigate the adverse effects of inactivity and the natural decline in functional capacity that occurs with aging \citep{nhanes1, nhanes2}. Furthermore, given that physical inactivity is a modifiable risk factor for various chronic diseases, including heart disease, diabetes, and specific types of cancer, the implementation of effective physical activity policies has the potential to extend the lifespan of this at-risk population, enhance their quality of life, and significantly reduce healthcare costs while alleviating the burden on the medical infrastructure.
	
	% This paper discusses the technical challenges of analyzing time series of physical activity derived from accelerometers, including variations in time series lengths, monitoring under free-living conditions, and missing patterns in the biological time series  \citep{10.1093/jrsssc/qlad007}. To address these challenges, we employ a methodology that summarizes time series information into multiple scalar metrics and incorporates the temporal dimension in metric estimation. Table \ref{table:tabla1} contains the variables considered in the analysis.

	The objectives of our analysis are twofold. First, from the list of variables in Table \ref{table:tabla1}- containing multiple physical activity summaries derived from accelerometers-  and other patient characteristics, we aim to identify a reduced subset of variables that maximizes predictive capacity. Second, we assess the variable selection capabilities of our approach and evaluate the efficacy of our predictive algorithm by comparing it with the traditional Accelerated Failure Time (AFT) log-normal model, assessing the capacity of our prediction model in creating time-to-event risk scores.
	
\begin{figure}[h!]
	\centering
	\includegraphics[width=0.7\linewidth]{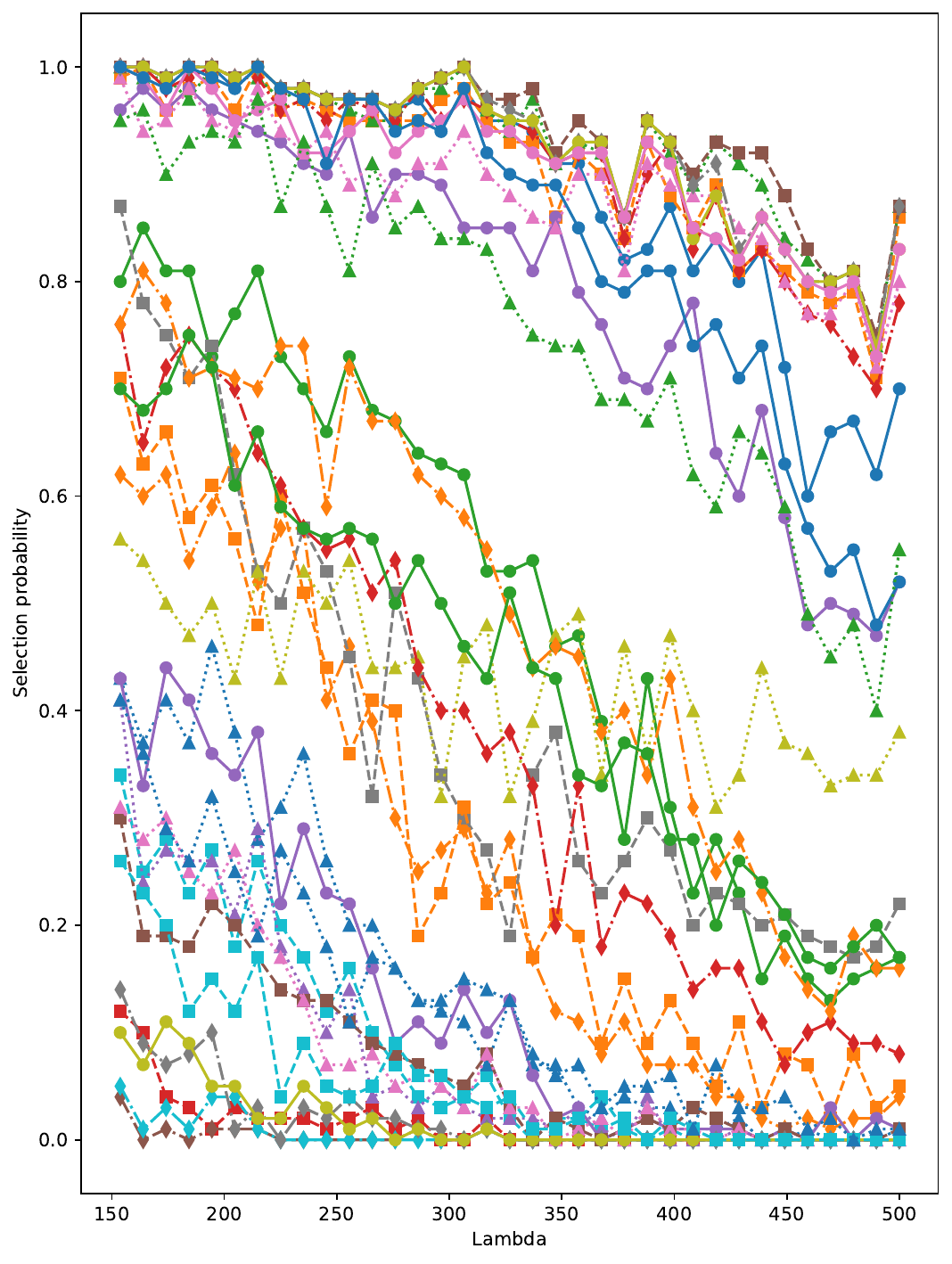}
	\caption{The stability selection paths for NHANES data. Each color represents a different covariate in Table \ref{table:tabla1}.The expected number $V$ of falsely selected variables is bounded by
		$
		E(V) \leq \frac{1}{2 \pi_{t h r}-1} \frac{q_{\Lambda}^2}{p}
		$. Setting $q_{\Lambda}$- the expected number of selected variables to $10$ and $\pi_{t h r}$ to $0.95$ we have $E(V) \leq 4$.}
	\label{fig:sub2}
\end{figure}

	To evaluate these objectives, we randomly divide the sample into a training set (80\% of the total) and a test set (the remaining 20\%) and run both our approach and the benchmark model using the variables listed in Table \ref{table:tabla1}. In our proposal, we further split the 80\% training set into two 40\% sets. The first part of the training sample is used for stability selection, and a low-dimensional fit is performed using the selected predictors in the second part of the sample. The performance of both models is evaluated using the concordance index on the remaining 20\% of the sample. Table \ref{table:tabla1} displays the variables selected by our model, and the estimated stability paths are shown in Figure \ref{fig:sub2}.
	
\begin{figure}[h!] 
	\centering

	\centering
	\begin{minipage}[b]{0.4\linewidth}
		\centering
		\includegraphics[width=\linewidth]{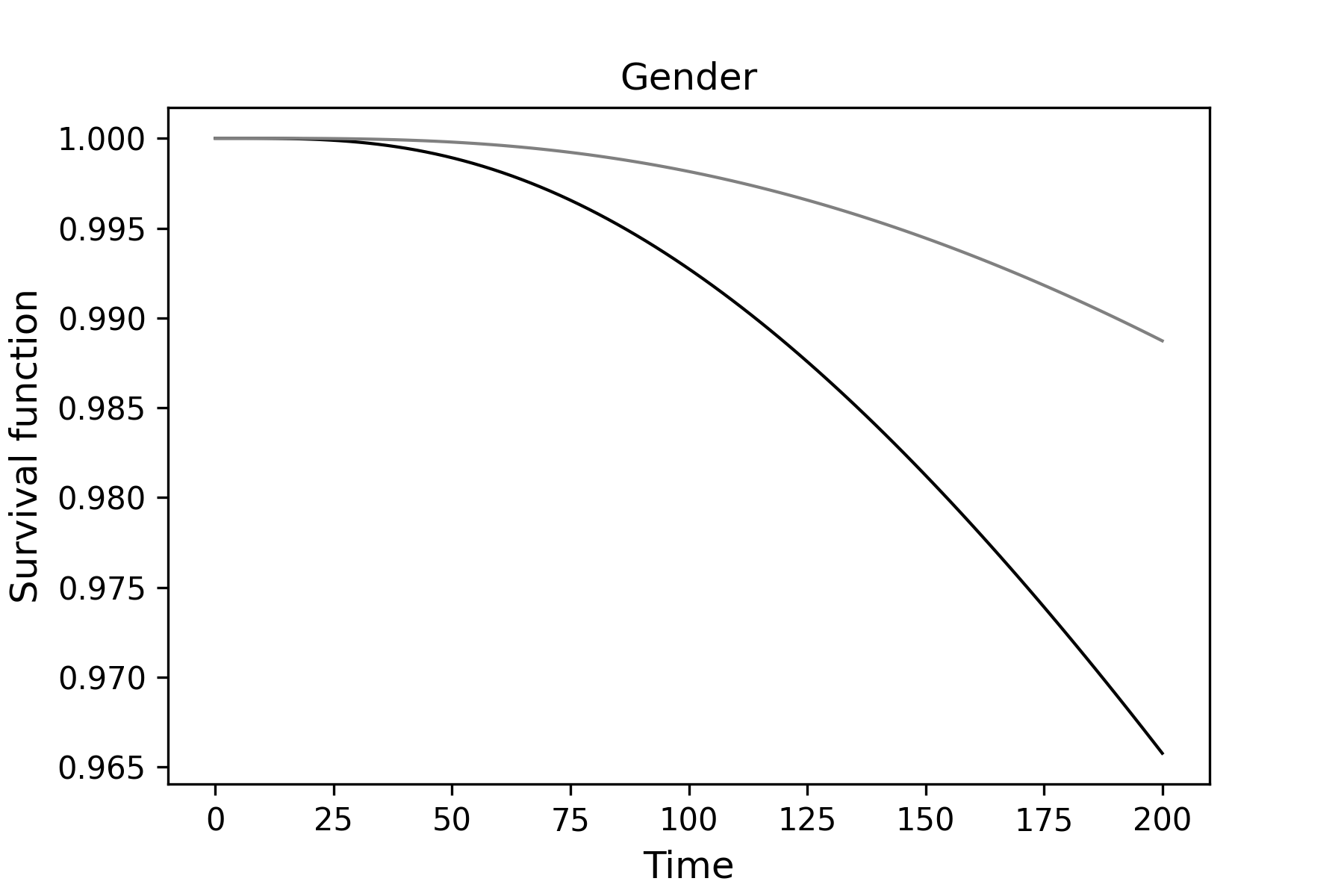}
		
	\end{minipage}%
	\hfill
	\begin{minipage}[b]{0.4\linewidth}
		\centering
		\includegraphics[width=\linewidth]{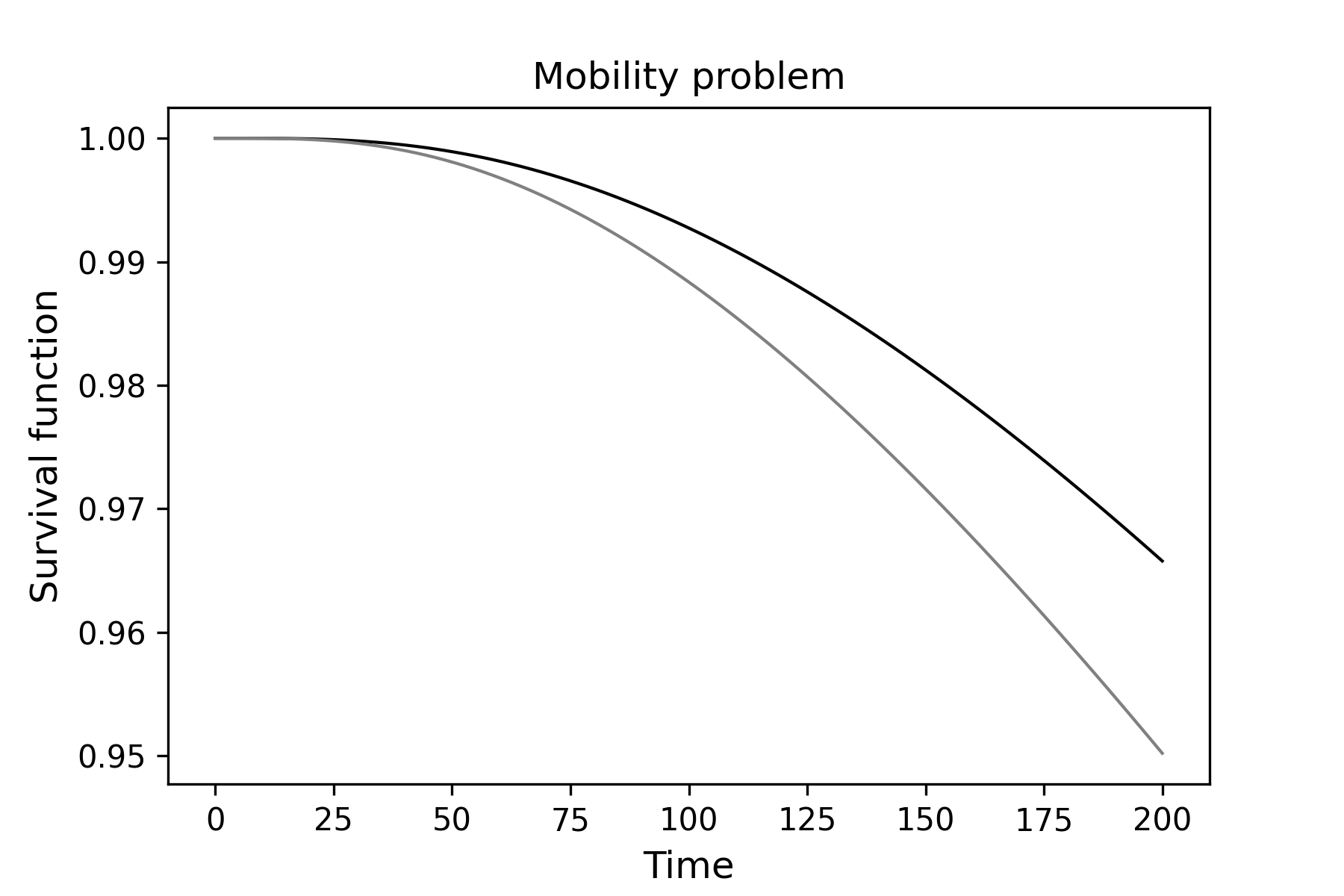}
		
	\end{minipage}
	
	\vspace{4ex}
	
	\begin{minipage}[b]{0.4\linewidth}
		\centering
		\includegraphics[width=\linewidth]{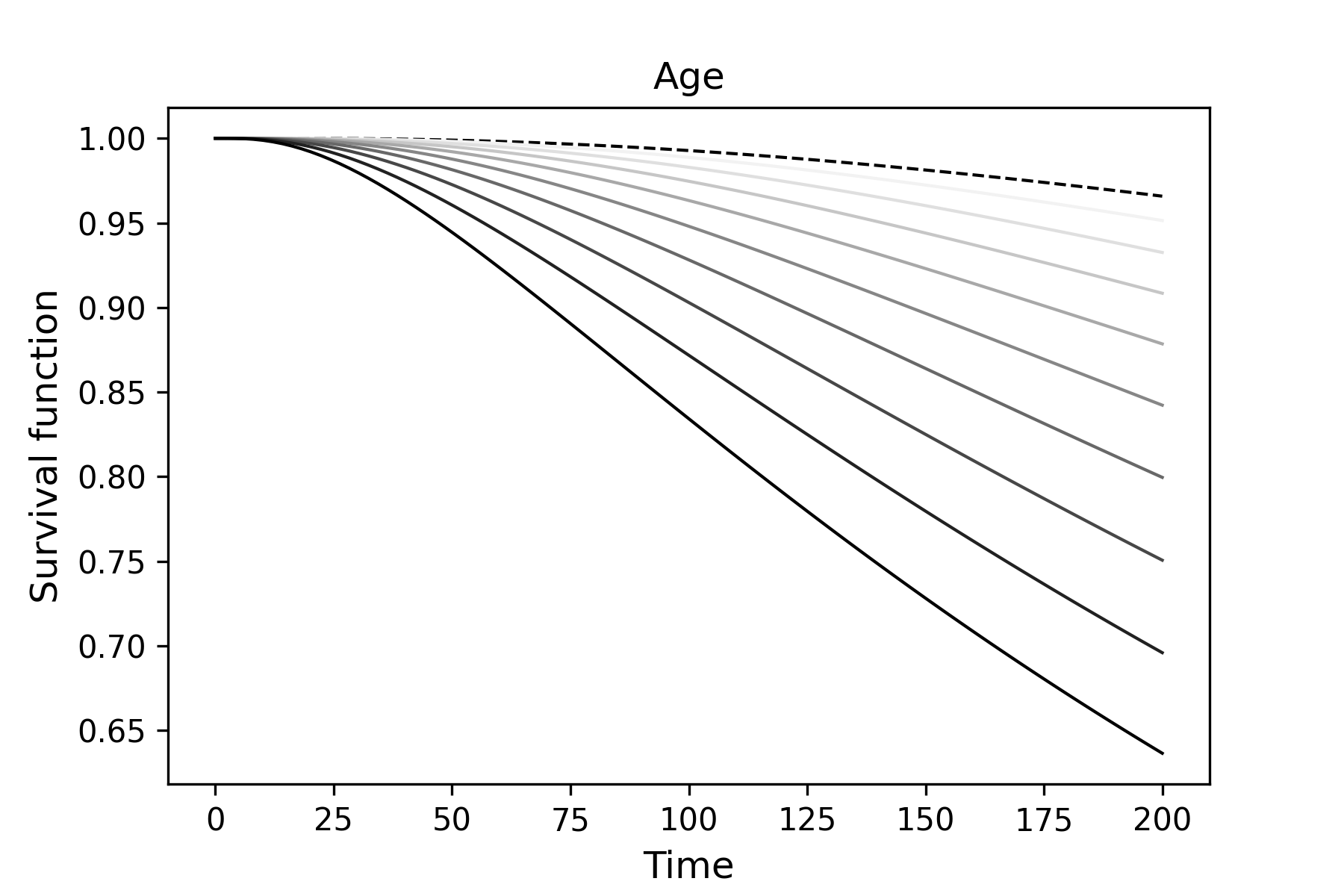}
		
	\end{minipage}%
	\hfill
	\begin{minipage}[b]{0.4\linewidth}
		\centering
		\includegraphics[width=\linewidth]{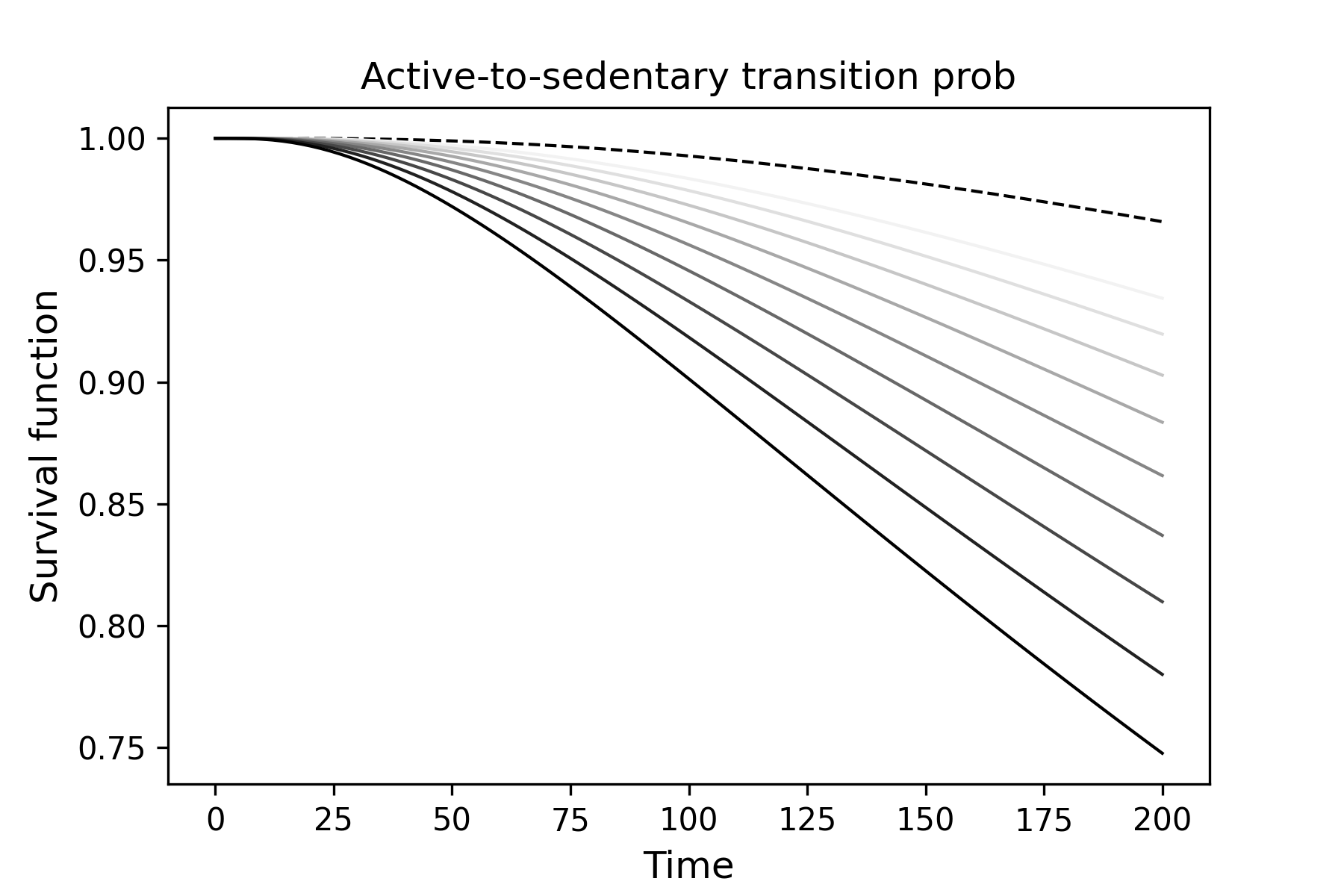}
		
	\end{minipage}
	\caption{Estimated survival functions for several configurations of the covariates (time scale in months). Note that the dashed graph, the baseline survival function, is the same across all the plots.  In (c), the different curves were estimated by doing forward passes of different ages through our model keeping the rest of the variables at zero. More intense greys correspond to higher ages, and the darkest solid grey corresponds to the age of the oldest person in the dataset. The same idea holds for (d), but varying ASTP and setting the rest of the covariates to zero. In (a), grey stands for female and black for male. In (b), grey stands for suffering from a mobility problem and black stands for the opposite. First hints of non-linearity can be perceived in that the grid of different age values on which predictions of the corresponding survival curves were launched was equally-spaced, whereas the estimated curves are not. }
	\label{curvas}
\end{figure}

	Our results are highly relevant in the domain of physical activity. Among the variables introduced in Table \ref{table:tabla1}, which include basic patient characteristics and a range of physical activity summaries, only the moderate to vigorous activity variable and some summaries of physical activity during the central time of the day are relevant in identifying patients with lower survival. These results align with the existing scientific literature, emphasizing the importance of moderate-to-vigorous physical activity \citep{stamatakis2022association}  and the significance of monitoring physical activity to identify increased risks across various diseases due to inactivity and transitions between inactivity and activity during the day \citep{schrack2019active}. Figure \ref{curvas} illustrates the survival curves predicted by our model for variables such as gender, age, mobility problem, and active-to-sedentary transition probability when fixing the rest of the covariates to zero. Notably, non-constant spacing of these conditional curves between consecutive values of the covariates reveals non-linearity in their dependence with survival. 
	
	In conclusion, our approach yields a higher concordance index of $0.79$ compared to the traditional linear log-normal AFT model, which produces a concordance index of $0.75$. These findings demonstrate the superior performance of our approach in predicting survival outcomes while comprehensively selecting the most relevant variables. Moreover, the use of variable selection algorithms allows us to identify the most relevant physical activity summaries in predicting patient survival, a topic of significant interest. As a next step, we suggest extending our algorithm to introduce modern distributional representations of accelerometer data as functional covariates, thereby addressing some limitations associated with using summaries \citep{matabuena2023distributional}.

%	\bibliography{citas}
	\bibliographystyle{apalike}

	\begin{table}[h]
		\centering
		\begin{minipage}[b]{\linewidth}
			\centering
			\caption{Performance of our proposal averaged across $B=500$ simulations}
			\label{tab:example1}
			\begin{tabular}{|c|c|c|c|}
				\hline
				\multicolumn{4}{|c|}{Example 1} \\
				\hline
				$n$ & c-index & $R^2$ & $\hat \sigma$ \\
				\hline
				500 & 0.8118 & 0.9647 & 0.3812 \\
				1000 & 0.8165 & 0.9776 & 0.4527 \\\
				2000 & 0.8198 & 0.9857 & 0.4732 \\
				\hline
				\multicolumn{4}{|c|}{Example 2} \\
				\hline
				$n$ & c-index & $R^2$ & $\hat \sigma$ \\
				\hline
				500 & 0.8110 & 0.9499 & 0.4139 \\
				1000 & 0.8156 & 0.9670 & 0.4596 \\
				2000 & 0.8192 & 0.9773 & 0.4868 \\
				\hline
			\end{tabular}
		\end{minipage}
		\\
		\begin{minipage}[b]{\linewidth}
			\centering
			\caption{Performance of the competitor averaged across $B=500$ simulations}
			\label{table:tabla2}
			\begin{tabular}{|c|c|c|c|}
				\hline
				\multicolumn{4}{|c|}{Example 1} \\
				\hline
				$n$ & c-index & $R^2$ & $\hat \sigma$ \\
				\hline
				500 & 0.6784 & 0.5664 & 1.8120 \\
				1000 & 0.6808 & 0.5659 & 1.8405 \\
				2000 & 0.6854 & 0.5702 & 1.8428 \\
				\hline
				\multicolumn{4}{|c|}{Example 2} \\
				\hline
				$n$ & c-index & $R^2$ & $\hat \sigma$ \\
				\hline
				500 & 0.8142 & 0.9173 & 0.8960 \\
				1000 & 0.8144 & 0.9180 & 0.9129 \\
				2000 & 0.8152 & 0.9179 & 0.9320\\
				\hline
			\end{tabular}
		\end{minipage}
	\end{table}
	
	\clearpage
	
\onecolumn
\begin{table}[h!]
	\small
	\caption{ Variable characteristics in the NHANES data.}\label{table:tabla1}
	\scalebox{1}{\begin{tabular}{|l|l|l|l|l|l|}
			\hline
			Variable & NHANES code & Description & Support & Selected   \\ 
			\hline
			$ 1$	&	\texttt{Gender}   &  & 0: Male, 1: Female& yes\\ 
			$ 2$	&	\texttt{Cancer}  & &  0: no, 1: yes& yes  \\ 
			$ 3$	&	\texttt{Stroke}  &  & 0: no, 1: yes & yes  \\

			$ 4$	&	\texttt{Diabetes}  & &  0: no, 1: yes & yes\\ 			
			
			$ 5$	&	\texttt{BMI}   & &  0: Normal, 1: Overweight & yes \\

			$ 6$	&	\texttt{CHF}  & Congesive Heart Failure& 0: no, 1: yes  & yes 
			\\ 
			
			$ 7$	&	\texttt{CHD}  & Coronary Heart Disease & 0: no, 1: yes  & yes
			\\

			$ 8$	&\texttt{Mobility Problem} & & 0: no, 1: yes  & yes \\ 
			
			$ 9$	&	\texttt{RIDAGEYR}  &Age at the interview time  &Continuous  & yes  \\ 
			
			$ {10}$	&	\texttt{LBXTC}   & Total Cholesterol& Continuous & no \\

			$ {11}$	&	\texttt{LBDHDD}  & Direct HDL-Cholesterol & Continuous &  no  \\
			
			$ {12}$	&	\texttt{SYS}  &Systolic Blood Pressure & Continuous   &  no\\ 
			
			$ {13}$	&	\texttt{TAC}   &Total volume of physical activity & Continuous   & no\\ 
			
			$ {14}$	&	\texttt{TLAC}   &Total log activity counts & Continuous   &  no \\ 
			
			$ {15}$	&	\texttt{WT}   &Accelerometer wear time & Continuous   &  no\\ 
			
			$ {16}$	&	\texttt{ST}   &sedentary Time & Continuous   &  no\\ 
			
			$ {17}$	&	\texttt{MVPA}   & Moderate‐to‐Vigorous Physical Activity& Continuous & yes \\ 
			
			$ {18}$	&	\texttt{ABout}   & & Continuous   &  no \\ 
			
			$ {19}$	&	\texttt{SBout}  & &Continuous    &  no\\ 
			$ {20}$	&	\texttt{SATP}   &  Sedentary-to-active transition prob.&Continuous  & no  \\ 
			
			$ {21}$	&	\texttt{ASTP}   & Active-to-sedentary transition prob.& Continuous   & yes  \\ 
			
			$ {22}$	&	$\texttt{TLAC}_1$  &Total log activity counts 12AM-2AM & Continuous   & no   \\

			$ {23}$	&	$\texttt{TLAC}_2$   & Total log activity counts 2AM-4PM& Continuous   & yes \\ 
			
			$ {24}$	&	$\texttt{TLAC}_3$   &Total log activity counts 4PM-6PM &Continuous    & no
			\\ 
			
			$ {25}$	&	$\texttt{TLAC}_4$  &Total log activity counts 6PM-8PM & Continuous   & no 
			\\

			$ {26}$	&	$\texttt{TLAC}_5$  &Total log activity counts 8PM-10PM & Continuous   & no  \\ 
			
			$ {27}$	&	$\texttt{TLAC}_6$   &Total log activity counts 10PM-12PM &Continuous    & no \\ 
			
			$ {28}$	&	$\texttt{TLAC}_7$   & Total log activity counts 12PM-2AM&Continuous    & no\\ 
			
			$ {29}$	&	$\texttt{TLAC}_8$  & Total log activity counts 2AM-4AM& Continuous   & no \\ 
			
			$ {30}$	&	$\texttt{TLAC}_9$  &Total log activity counts 4AM-6AM &Continuous    & no  \\ 
			
			$ {31}$	&	$\texttt{TLAC}_{10}$  & Total log activity counts 6AM-8AM& Continuous   & no \\

			$ {32}$	&	$\texttt{TLAC}_{11}$ &Total log activity counts 8AM-10AM & Continuous   & no\\ 
			
			$ {33}$	&	$\texttt{TLAC}_{12}$  & Total log activity counts 10AM-12AM& Continuous   & no \\ 
			\hline

	\end{tabular}}	
\end{table}

\end{document}